# Supervised Learning and Anti-learning of Colorectal Cancer Classes and Survival Rates from Cellular Biology Parameters

Chris Roadknight[a], Uwe Aickelin[a], Guoping Qiu[a], John Scholefield[b] and Lindy Durrant[b]

a. Intelligent Modelling & Analysis Research Group (IMA)
School of Computer Science
b. Faculty of Medicine & Health Sciences
The University of Nottingham
cmr,uxa,qiu@cs.nott.ac.uk

*Abstract*— In this paper, we describe a dataset relating to cellular and physical conditions of patients who are operated upon to remove colorectal tumours. This data provides a unique insight into immunological status at the point of tumour removal, tumour classification and post-operative survival. Attempts are made to learn relationships between attributes (physical and immunological) and the resulting tumour stage and survival. Results for conventional machine learning approaches can be considered poor, especially for predicting tumour stages for the most important types of cancer. This poor performance is further investigated and compared with a synthetic, dataset based on the logical exclusive-OR function and it is shown that there is a significant level of "anti-learning" present in all supervised methods used and this can be explained by the highly dimensional, complex and sparsely representative dataset. For predicting the stage of cancer from the immunological attributes, anti-learning approaches outperform a range of popular algorithms

## I. INTRODUCTION

Colorectal cancer is the third most commonly diagnosed cancer in the world. Colorectal cancers start in the lining of the bowel and grow into the muscle layers underneath then through the bowel wall [11]. TNM staging involves the Classification of Malignant Tumours

- Tumour (T). Size of the tumor and whether it has invaded nearby tissue
- Nodes (N). The extent to which regional lymph nodes involved
- Metastasis (M). This is the spread of a disease from one organ or part to another non-adjacent organ.

4 TNM stages (I,II,III,IV) are generated by combining these three indicator levels and are allied with increasing severity and decreasing survival rates.

Treatment options include minor/major surgery, chemotherapy, radiotherapy but the correct treatment is heavily dependent on the unique features of the tumour which are summarised by the TNM staging. Choosing the correct treatment at this stage is crucial to both the patient's survival and quality of life. A major goal of this research is to automatically optimize the treatment plan based on the existing data.

The data for this research was gathered by scientists and clinicians at the University of , Nottingham. The dataset we use here is made up of the 84 attributes for 462 patients. The attributes are generated by recording metrics at the time of tumour removal, these include:

- Physical data (age, sex etc)
- Immunological data (levels of various T Cell subsets)
- Biochemical data (levels of certain proteins)
- Retrospective data (post-operative survival statistics)
- Clinical data (Tumour location, size etc).

The goal of this research is two-fold, we hoped to be able to use the attribute set to accurately predict:

- The TNM stage assigned by the clinical team.
- The subsequent survival of the patient

We show in this paper that both of these tasks are extremely difficult using conventional techniques and that the dataset might belong to a subset of dataset that require a unique approach.

## II. PRE-PROCESSING

The dataset supplied is a biological dataset and as such has a rich complement of preprocessing issues. 11.32% of the values are missing, with some attributes having over 40% missing values and some patients having over 30% missing values.

Missing data poses a problem for most modelling techniques. One approach would be to remove every patient or every attribute with any missing data. This would remove a large number of entries, some of which only have a few missing values that are possibly insignificant. Another approach is to average the existing values for each attribute and to insert an average into the missing value space. The appropriate average may be the mean, median or mode depending on the profile of the data.

Much of the data takes the form of human analysis of biopsy samples stained for various markers. Rather than raw cell counts or measurements of protein levels we are presented with threshold values. For instance, CD16 is found on the surface of different types of cells such as natural killer, neutrophils, monocytes and macrophages. The data contains a simple 0 or 1 for this rather than a count of the number of cells. This kind of manual inspection and simplification is true for most of the data and any modeling solution must work with this limitation.

It is apparent that there are some existing strong correlations in the data. By using a combination of correlation coefficients and expert knowledge the data was reduced down to a set of ~50 attributes. This included removing several measurements that were hindsight dependent (ie. chemo or radio treatment) and correlated with TNM stage. (ie. Dukes stage).

Single attribute relationships exist within the dataset but are not strong. Analysis of single attributes can yield a greater than 65% prediction rate when attempting to predict which TNM stage a patient was classified as but only ~55% when the TNM stages were restricted to the more interesting (TNM stage 2 or 3). If we look at CD59a and CD46 threshold values we can see that they are loosely related to survival (figure 1) with elevated levels of each indicating a reduction in survival averaging ~13 (Figure 1a) and 6 months. (Fig. 1b) yet neither are a strong discriminator of TNM stage 2 or 3 tumours.

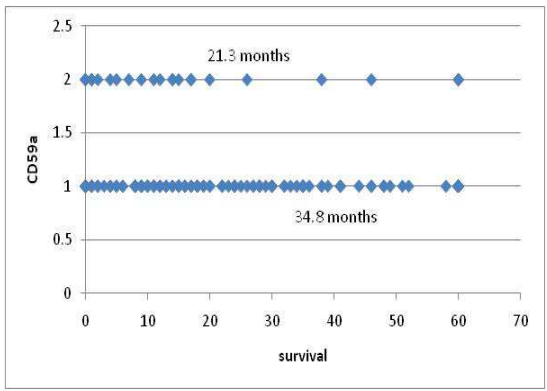

Figure 1a. Relationship of CD59a to survival with average survival rates.

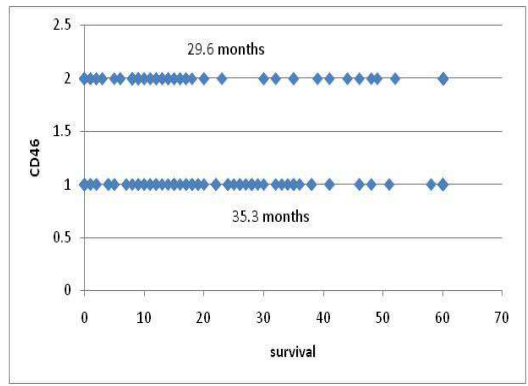

Figure 1b. Relationship of CD46 to survival with average survival rates.

### III. LEARNING

It is relatively trivial to build a model that best fits the data, even with numerous attributes and missing values. Unfortunately this model is very likely to be memorising unique combinations of values for each patient. This is why models are tested on an unseen test set to decide how well the trained model generalises to the "rest of the world".

Define abbreviations and acronyms the first time they are used in the text, even after they have been defined in the abstract. Abbreviations such as IEEE, SI, MKS, CGS, sc, dc, and rms do not have to be defined. Do not use abbreviations in the title or heads unless they are unavoidable.

#### A. TNM Stage.

Several methods were used in an attempt to predict the appropriate TNM stage of a patient from their attribute set. The methods used included Bayesian Networks [7], Naïve Bayes Classifier [8], CART [3], Multilayer Perceptron [4] and SVM [9]. These were either self-programmed, available in the WEKA toolkit [6] or used other existing software suites [5].

When initially looking at all 4 TNM stages there was some success at predicting stages from the attribute set, particularly when some of the patients and attributes with the most missing data were removed. Most success was achieved when predicting TNM stage 1 and 4, which were the least and most severe stages respectively. A Multilayer Perceptron (MLP) was trained using back-propagation of error. This artificial neural network architecture included 5 sigmoid transfer function hidden units and a linear transfer function output unit. The desired output for the for TNM stage 1,2,3 and 4 were rescaled to 0.2, 0.4, 0,6 and 0.8 to allow for efficient separation. This approach showed excellent accuracy on the training set (Fig 2a) and showed some promise at predicting TNM stages 1 and 4 for the unseen test set (Figure 2b) but was clearly very poor at predicting stages 2 and 3. This method of graded linear output makes for a neural network of low complexity but assumes a linear progression through the TNM stages. An approach using 4 independent binary outputs performed equally poorly.

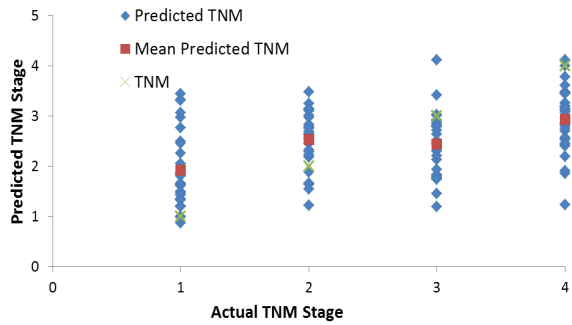

Figure 2a. Neural Network Prediction of TNM stage for training set

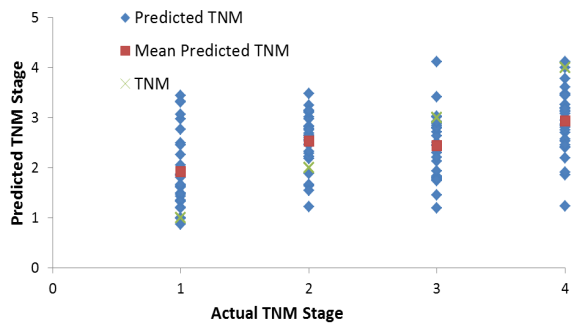

Figure 2b. Neural Network Prediction of TNM stage for an unseen test set

Next a CART approach was used, Classification and regression trees (CART) are a non-parametric decision tree learning technique that produces either classification or regression trees. Decision trees are formed by a collection of rules based on attributes in the dataset based on rules that achieve the best division to differentiate observations based on the dependent variable. This recursive process continues until pre-defined stopping rules are met. Each branch of the tree ends in a terminal node. Each observation falls into exactly one terminal node, and each terminal node is uniquely defined by a set of rules. A CART approach achieved similar results when looking at all 4 TNM stages but slightly better results could be achieved by just looking at TNM stages 2, 3 and 4. (Figure 3a) with particular importance being assigned to cleaved caspase 3 (CC3) proteins, a sample CART tree is shown in figure 3. CC3 has been shown to play an important role in tumour apoptosis [10].

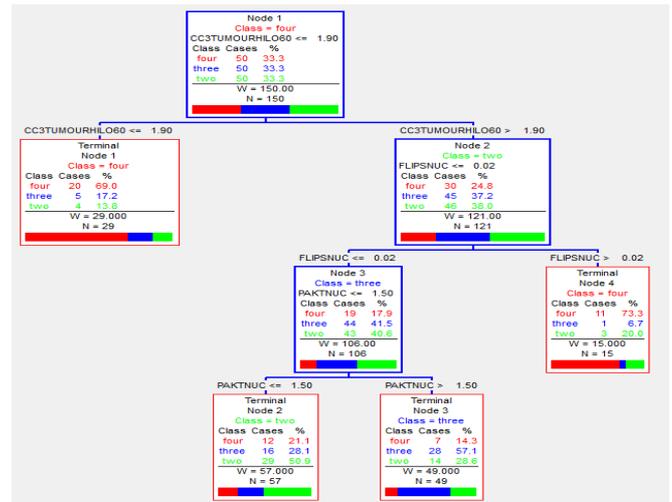

Figure 3. CART tree for prediction on TNM class

Differentiating between TNM stage 2 and 3 is a priority for this research, these are the stages where correct choice of post-operative treatment are most important. We trialed all 5 algorithms (SVM, Bayesian Network, Artificial Neural Network, Naïve Bayes Classifier and CART) on 3 different attribute sets (55, 45, 35) and used 3 different testing regimes (10 fold cross validation, 5 fold cross validation and a 33% random selection). Table 1 shows the results from this, the only approach that performed better than guessing was Naive Bayes and this was only on one of the three attribute sets. Furthermore, if we look at the performance of all approaches for each validation technique as a whole it can be said that the approaches performed significantly worse than guessing. It must be remembered that algorithms were optimised for test set performance and performance of the training set was much better (80-98% accurate). This exceptionally poor performance will be discussed more in section 4.

Table 1. Test Set performance of different algorithms

|  | 10 fold cross-validation | | | |
| --- | --- | --- | --- | --- |
|  | 55 Attributes | 45 Attributes | 35 Attributes |  |
| Naive Bayes | 52.43 | 48.26 | 41.46 | Mean |
| ANN | 45.83 | 45.35 | 46.34 | 46.67 |
| CART | 44.79 | 48.25 | 45.73 | StDev |
| SVM | 44.44 | 45.93 | 44.51 | 2.65 |
| BayesNet | 48.61 | 49.41 | 48.78 |  |
|  | 5 fold cross-validation | | | |
| Naive Bayes | 51.39 | 41.86 | 40.85 | Mean |
| ANN | 46.53 | 44.77 | 44.51 | 45.69 |
| CART | 44.1 | 50 | 45.122 | StDev |
| SVM | 48.61 | 43.6 | 39.63 | 3.48 |
| BayesNet | 49.65 | 45.93 | 48.78 |  |
|  | Random 33% | | | |
| Naive Bayes | 56.12 | 43.1 | 42.86 | Mean |
| ANN | 45.92 | 34.48 | 41.07 | 42.78 |
| CART | 50 | 39.66 | 39.28 | StDev |
| SVM | 40.82 | 27.58 | 44.64 | 6.83 |
| BayesNet | 46.93 | 50 | 39.29 |  |

## B. Survival

Several of the attributes presented in the dataset pertain to the survival of the patients after their operation to remove the tumour. The number of months the patient has survived, whether they are still alive or not and how they died (if dead) are all available. Figure 4 shows survival curves for patients with greater than 60 months survival or those that died of colorectal cancer prior to the 60 month period. The strong difference between survival rates in TNM stage 1 and 4 patients is apparent (ie. at 30 month the survival rate is approximately 95 and 5%). The difference between patients with TNM stage 2 and 3 cancers is less apparent, for 18 months there is very little difference between mortality for TNM stage 2 and 3 patients.. After 30 months deaths from colorectal cancer for TNM stage 1 patients increase quite quickly, in percentage terms steeper than any other TNM class.

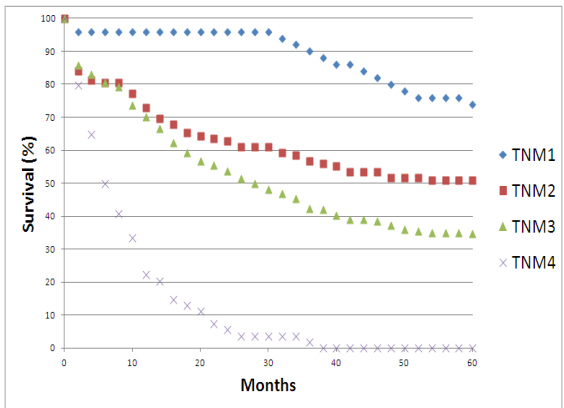

Figure 4. Survival Curves for patients at all 4 TNM stages

Again focusing on just TNM stage 2 and 3 patients we attempted predict survival at different thresholds using AI techniques, this time Naïve Bayes, ANN and CART. The term "survival" is obviously somewhat subjective so we used several time periods to represent survival ranging from 12 to 60 months, if a patient survived for the assigned number of months they were deemed to have survived.

If we take just TNM stage 2 and 3 patients again, figure 5 shows how well three techniques predicted survival for an unseen test set, the average of all three techniques is also shown. It is apparent that these techniques could perform slightly better than guessing at all survival thresholds but with average performances of between 55 and 60% on an unseen test set (using 10 fold cross validation) the performance is far from impressive suggesting the issues discussed in section 3.1 are still present. With CART performing worse than guessing at a high survival threshold and an ANN performing worse than guessing at a low survival threshold.

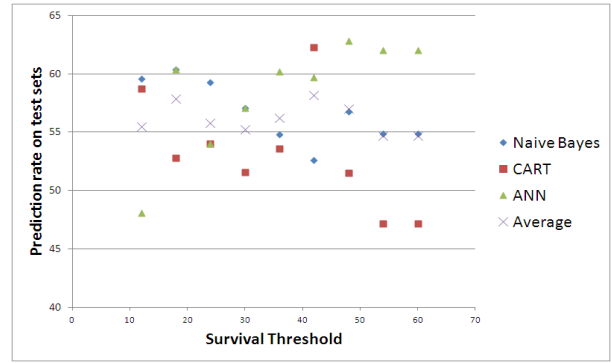

Figure 5. Prediction accuracy when modeling TNM stages 2 or 3

## IV. ANTI-LEARNING

In many cases the results shown in section 3 show some very poor prediction on an unseen test set, sometimes below what would be expected for random guessing. This kind of behavior is rare but when it has been observed one of the dataset types it has been observed in is biological data in general and cancer data in particular. We investigated this further by running a full range of predictive techniques on several pre-processed versions of the original dataset with several correlation techniques. The results from this were tabulated in Table 1. It is apparent that in all cases results on a test set are a small but significant amount below 50%, which would be the value for a random selection. Results were even lower when hybrid techniques such as Bagging [12] and Boosting [13] were used. If we assume anti-learning is present and invert the outcome of the model we find superior prediction results to any learning approach tested. Table 2 shows a comparison of the best performing learning and anti-learning techniques for 3 processed versions of the dataset with 55, 45 and 35 attributes, with the anti-learning results showing up as better on all 3 datasets.

Synthetic and real world datasets have been shown to express similar anti-learning properties. The simplest example being the exclusive-OR (XOR) problem, which can be summarised as a logical function on two operands that results in a value of true if exactly one of the operands has a value of true. This can be tabulated and plotted as follows, with X and Y being the two operands and Z being the result (figure 6). An exclusive-AND function is just the opposite where two operands that results in a value of true if neither or both of the operands are true.

Table 2. Comparison of Learning and Anti-learning methods of predicting TNM stage for datasets with 35, 45 and 55 attributes

|  | Dataset Attributes | | |
| --- | --- | --- | --- |
| **Method** | **55 Attributes** | **45 Attributes** | **35 Attributes** |
| Learning | 52.43% (Naive Bayes) | 49.41% (BayesNet) | 46.34% (ANN) |
| Antileaning | 55.56% (SVM) | 54.65% (ANN) | *58.54%* (Naive Bayes) |
| Anti-learning + Boosting | *57.25%* (SVM) | *56.9%* (ANN) | *58.54%* (Naive Bayes) |

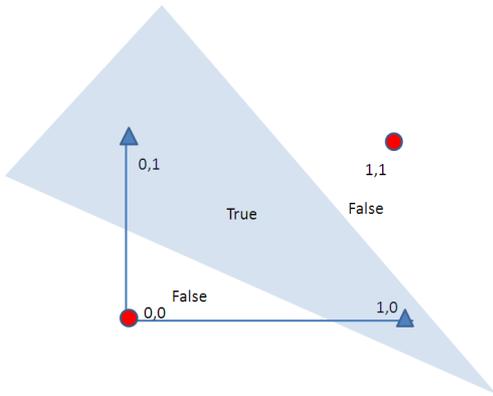

Figure 6. Graph and Table showing a standard 2 dimensional XOR dataset

This dataset can be learned by an AI approach capable of non-linear feature extraction, an example would be an artificial neural network, but only if all 4 operand couples are presented with the desired output. This leaves no samples for testing. If we present any 3 of the 4 examples to any machine learning approach they will generalise to a point where they predict the unseen test value wrong 100% of the time, in effect they will "anti-learn" the problem. This is a trivial, abstract example but is an important indication that if too few datapoints are presented to a machine learning solution it is possible that they will not only perform poorly (ie. equal to guessing) on an unseen test set but actually perform WORSE than guessing. When we are dealing with real world datasets with many attributes and relatively few samples, the possibility that the N dimensional search space is badly represented is distinct.

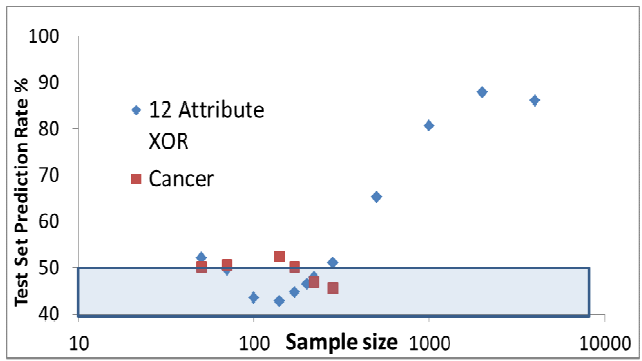

Figure 7 Predictive performance of Neural Network and increasing sample sizes

We can generate a 12 dimensional synthetic XOR style dataset by taking all possible combinations of a 12 attribute binary dataset (4096 combinations) and passing them through a series of XOR and XAND processes:

{ [ ( A xor B ) xor ( C xand D ) ] xor [ ( E xor F ) xand ( G xand H ) ] } xor { [ ( I xor J ) xor ( K xand L ) ] }

If we take this 12 dimensional exclusive-OR and exclusive-AND problem we can achieve degrees of anti-learning when small percentages of the total dataset are presented to a learning algorithm. Figure 7 shows how presenting most of the available data yields high test set performance (~90%) but reducing the sample size for training and testing reduces the test set performance to a point where it drops below 50%, reducing it even further means any prediction tends towards 50% (guessing). Initial results for a real world cancer dataset appear to occupy an area that would suggest the available data represents only a small sample of a much bigger, non-linear 'complete' set. Another approach to show anti-learning exists in this dataset and how it differs from overtraining or the absence of any learnable features is to show the difference between training and testing for a dataset over a range of modeling configurations, in this case ANN architecture. If we take a very simple, single hidden unit ANN we can achieve slightly higher than guessing performance for the Colorectal Cancer training dataset (figure 8), but below guessing for test set. As we increase the number of hidden units we see an increase in training set performance up to nearly 100% at 20 hidden units but performance on an unseen test set yields a less than 50% performance in all cases. In cases of a normal, learnable dataset we can see that the test set performance increases up to an optimal number of hidden units of 7 before overtraining occurs and the test set performance decreases. If the dataset just consists of random numbers the test set performance approximates to 50% which is the same as guessing as there are no general features in the data but adding hidden units allows a degree of memorization in the training set. From these results we can conclude that the cancer dataset consists of a mixture of unlearnable attribute relationships and anti-learning relationships.

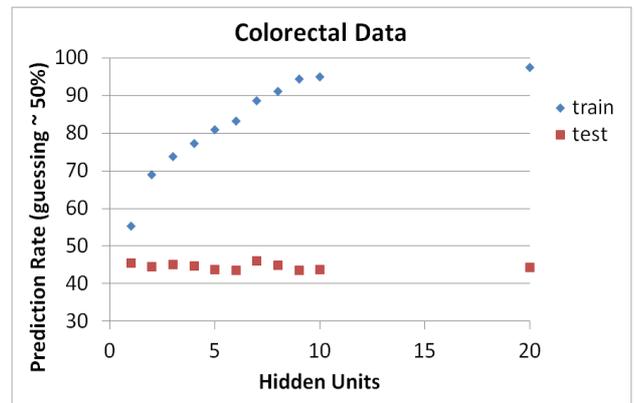

Figure 8. Training and test set performance for colorectal cancer data as the number of hidden units is increased

## V. CONCLUSIONS

We have presented results for a unique dataset based on the biochemical and factors associated with colorectal tumour patients. This dataset is limited in many ways, but extremely important nonetheless and modeling any relationships or features based on the dataset to hand is an urgent priority. Generally, whether attempting to predict TNM stages or survival, patients at TNM stage 1 and 4 have more clear indicators in the attribute set. TNM stage 2 and 3 provides a much more challenging prediction task, so much so that the TNM stage appears much less important when predicting survival for these 2 stages than other indicators.

Rule tree, Bayesian and Neural approaches have been used with some limited success for prediction, but in most experiments there is a lack of repeatable success in developing a model that accurately predicts survival or TNM stage on an unseen test set. One possible reason for this could be overfitting, though a well-constructed ANN or CART tree shouldn't exhibit overfitting and in any case they shouldn't be WORSE than guessing. Another possibility is poor or inaccurate labeling of patients tumour stages. But again this should only result in poor performance on the unseen test set. Modelling a dataset using all available data may produce the best possible model if the modelling process is ideally carried out, but ideal modelling is much more difficult without a test set. Methods such as Correlated Activity Pruning [14] may be useful in ensuring a minimal sized model and will be one focus of future research. There might be improvement to learning by using recent advancements such as multiple kernel learning [15] but it is just as likely, as with boosting, methods that improve learning may be just as effective at improving anti-learning.

This failure to accurately classify TNM stages or survival periods may in fact be useful if we suspect there are subsets within the groups. The failure to correctly classify a set of patients may mean these patients have different characteristics while still expressing the same classification of tumour. This would imply that treatment based solely on tumour classification would be sub-optimal.

Overall this is an iterative process with a large number of steps, each providing more insight into the dataset and its modelling. We are still at the stage where we are filtering and focusing the original data so that we arrive at the most important, complete dataset for modelling the relationship between tumour markers/immunology, tumour stage and survivability. It is also very significant that preprocessing the patient's data (selecting based on different thresholds) has a significant difference on the resulting models.

We have proposed an explanation for the results which is a phenomenon called "Anti-learning". Here, unique characteristics of the dataset lead to a condition where validation on an unseen test set produces results significantly and repeatedly worse than guessing. Interestingly, one real world dataset that demonstrates this behaviour is very similar to the dataset used here, being the classification of response to chemoradiotherapy in Esophageal Adenocarcinoma patients using microarray data of biopsied patients [1][2]. Work with a 12 dimensional exclusive-OR problem shows that when only a small portion of the dataset is available there is a real possibility that anti-learning will be present. It is possible to then infer that with some highly dimensional complex biological data sets, when we have a relatively small sample size, anti-learning may also exist. Initial experiments appeared to show that the best possible approach to classifying patients with TNM stage 2 and 3 tumours was to focus on anti-optimising the learning process to achieve the worse possible test set performance and then inverting the underlying model. Overall when looking for test set performance on the important TNM stage 2 and 3 patients, the best possible results can be achieved if we inverted the answer supplied by an ADABoosted SVM or ANN. Using this method it is possible to achieve reliable prediction rates of 55-60% on an unseen test set of higher than any learning algorithm attempted. Taking this approach involves a significant leap of faith but we have shown that this method is optimal when dealing with a small sample of a compounded exclusive-OR dataset. It is not impossible to imagine that many complex biological datasets also present us with a small, noisy sample of a much bigger complex dataset and this must be investigated further.

This dataset will be made available, in an anonomysed form for other research groups to apply their own methods to ascertain the true extent of the anti-learning behavior. Interested parties should contact the authors about this.